\documentclass{llncs}
\usepackage[numbers]{natbib}

\usepackage{mathtools} 
\usepackage{url}
\usepackage{amsmath} 
\RequirePackage[utf8]{inputenc} 
\usepackage{microtype} 
\usepackage{listings}
\def\linebreak{\raisebox{0ex}[0ex][0ex]{\ensuremath{\hookleftarrow}}}
\def\postlinebreak{\raisebox{0ex}[0ex][0ex]{\ensuremath{\hookrightarrow\space}}}
\newcommand*\DNumber{\addtocounter{lstnumber}{-1}} 
\lstdefinestyle{codeListing}{%
  basicstyle=\small,
  breaklines=true,
  showtabs=true,
  numbers=left,
  numberstyle=\small,
  numberblanklines=false,
  escapechar=\#,
  commentstyle=\emph,
  keywordstyle=\textbf,
  emphstyle=\textbf,
  prebreak=\linebreak,
  postbreak=\postlinebreak,
  breakatwhitespace=true,
  frame=L,
  emph={switch,case,for,if,emit,reduce,map,new},
  xleftmargin=\parindent,
  language=C,
  captionpos=b
}
\usepackage{csquotes} 
\usepackage{booktabs} 
\usepackage{hyperref}
\RequirePackage[hyphenbreaks]{breakurl} 
\usepackage{cleveref} 

\makeatletter
\providecommand\phantomcaption{\caption@refstepcounter\@captype}
\makeatother
\usepackage{fancyhdr}
\newcommand{\semweb}{Semantic Web}
\hypersetup{%
  pdftitle    = {Large-Scale Reasoning with OWL},
  pdfauthor   = {Michael Ruster},
  pdfkeywords = {large-scale reasoning, reasoning, semantic web, billions, triples, OWL 2 RL, OWL pD*, OWL 1, OWL 2, QueryPIE, WebPIE, MapReduce, query rewriting, materialisation, materialization, backward chaining, forward chaining, OBDA},
  colorlinks  = true,
  unicode     = true,
  linkcolor   = black,
  citecolor   = black,
  filecolor   = black,
  urlcolor    = black,
}
\title{Large-Scale Reasoning with OWL}
\author{Michael Ruster}
\institute{University of Koblenz-Landau, Campus Koblenz}
\begin{document}
\maketitle
\pagenumbering{arabic}

\section{Introduction}
In recent years, the \semweb{} has grown in size and importance.
More and more knowledge is stored in machine-readable formats like RDFS or OWL\@.
For many applications, knowledge extraction and reasoning is one of the core requirements.
Through reasoning, knowledge can be logically derived that is not explicitly present in the data.
Due to the complexity and amount of knowledge on the \semweb{}, this can easily become a difficult task.
The bottlenecks are the time used for processing a query as well as the memory needed while reasoning.

This paper outlines common approaches for efficient reasoning on large-scale data.
It therefore presents techniques implemented in reasoners, which are able to process billions ($10^9$) of triples.
The paper focuses on OWL because it is widely used as a knowledge representation ontology language on the \semweb{} and because it is rich in features.
First of all, a brief introduction to the \semweb{} is given in \autoref{s:semweb}.
For this, common properties of it are highlighted which may influence the choice of language selection for knowledge representation.
Subsequently, \autoref{s:owl} will give an overview of OWL and some of its sublanguages.
The choice of sublanguages is shortly explained based on the requirements of modelling knowledge on the \semweb{}.
\autoref{s:techniques} illustrates two main techniques for large-scale reasoning.
Differences between the approaches are being highlighted.
Furthermore, for both techniques, one example reasoner is presented together with some optimisation strategies they implement.
Finally, \autoref{s:conclusion} summarises this paper.

\section{Semantic Web and Language Features}\label{s:semweb}
This section focuses on the \semweb{} and some of its most important properties.
Thereby, it introduces the various important terms which are used throughout this paper.
Furthermore, this section describes two different logics that can be of use for modelling \semweb{} data.

The \semweb{}'s purpose is to extend the World Wide Web by encoding its information in machine-parseable ways~\citep{berners-lee_semantic_2001}.
As a result, it should be possible for machines to easily extract a Web page's \enquote{meaning}~\citep{shadbolt_semantic_2006}.
The \semweb{} is built using various technologies like \emph{ontologies} to achieve this.
Ontologies are conceptual models, encoding a set of terms and relationships between them~\citep{berners-lee_semantic_2001}.
This enables the organisation and exchange of information as well as reasoning on it.
Formal descriptions of ontologies are given through \emph{ontology languages}. 
Ontology languages are using logics to express knowledge. 
OWL is built around description logics to which an overview is given in \autoref{s:techniques}.
Its statements can be expressed through a subject-predicate-object structure which are called \emph{triples}.
All statements that describe the taxonomy of the domain by expressing terminological knowledge are called the \emph{TBox} of the ontology~\citep{baader_description_2003}. 
Likewise, all statements describing assertions about instances like their properties and relations to other instances are called the \emph{ABox}~\citep{baader_description_2003}. 

\semweb{} data may originate from many diverse fields such as biology, medicine~\citep{shadbolt_semantic_2006} or journalism like the New York Times\footnote{\url{http://data.nytimes.com/} --- last accessed 25 March 2015, 11:00}. 
Hence, the knowledge to be represented is heterogeneous and may require different statement capabilities of its ontology language.
A richer amount of features may also mean a higher complexity while reasoning~\citep[cf.][]{krotzsch_owl_2012}.
Nevertheless, depending on the domain to model, more features might be needed.
For example, generally, every statement is either true or false.
But modelling uncertainty or imprecision is then not possible e.g.\ in classical description logic, a statement cannot express that a paper is \enquote{almost} finished.
A solution for this is to introduce a finer range of truth values.
This can be done e.g.\ with \emph{fuzzy logic} by assigning such an additional truth value in the range of $[0,1]$ to each triple~\citep{zadeh_fuzzy_1965}. 

Similarly, reasoning over incomplete and conflicting knowledge can be problematic for instance when trying to calculate a transitive closure over multiple triples.
A common example expresses that birds can fly and that penguins are birds.
But penguins form an exception regarding the capability of flight while \emph{most} birds are actually able to fly.
\emph{Defeasible logic}~\citep{nute_defeasible_2003} allows expressing sentences like \enquote{generally, birds can fly} while also modelling that penguins cannot although they are birds \citep[cf.][]{lukasiewicz_expressive_2008}. 
This logic introduces three main elements~\citep{nute_defeasible_2003}:
\begin{description}
  \item[Strict rules] model knowledge that is true in all cases like \enquote{swallows are birds}.
  \item[Defeasible rules] express that something \enquote{typically} holds like \enquote{birds can fly}.
        They can be defeated by other defeasible or strict rules.
        Rules can be assigned priorities to determine which rule may defeat another one~\citep{nute_defeasible_2003,garcia_defeasible_2004}.
  \item[Undercutting defeaters] formulate possible exceptions to defeasible rules without them being expressive enough to allow any concrete inference.
        For example, \enquote{an injured bird \emph{might} not be able to fly} would not allow an inference that all injured birds are unable to fly.
        Instead, it highlights that there might be exceptions which would make an inference impossible.
\end{description}

Fuzzy logic and defeasible logic may seem similar at first but are indeed different.
As \citet{covington_defeasible_2000} notes, fuzzy logic allows reasoning with a certain level of uncertainty or imprecision. 
Whereas defeasible logic ignores these degrees of truths and instead expresses that some rules may be overridden by others.

Both embody interesting concepts for the \semweb{}.
Fuzzy logic can be used to express various level of trust in certain sources.
For example, an article by a renowned newspaper will often be more trustworthy than one by a rather unknown personal blogger.
Defeasible logic helps modelling conflicting knowledge in a different way.
In the example before, the information by the blogger can be defeated by the one published by the newspaper.
It remains unclear how much more credible the newspaper is compared to the blogger.
Yet, the information from the newspaper article can be seen as truth instead of only information closer to the actual truth.
This allows modelling of incomplete knowledge with knowledge capturing typical states and being defeated by more specific rules.
Hence, when knowledge changes would appear that are expressed through defeasible rules, these defeasible rules could immediately be added without entailing further processing~\citep{garcia_defeasible_2004}. 

It depends on the ontology developers whether they see a need for these logics to describe their data.
In some contexts, it can be useful.
However, it always adds complexity to the reasoning process.
In the next section, sublanguages of OWL are discussed to illustrate possible choices in expressiveness that are already provided by OWL without any further extensions.

\section{OWL and its Sublanguages}\label{s:owl}
This section gives a quick introduction to OWL and its sublanguages in terms of their features.
The purpose of this section is to illustrate that the choice of an ontology language is an important decision for reasoning on big data.
For this purpose, reasoning complexities of most languages are given.
Their most important properties are presented and thus it should be understood, why some languages are commonly used for this type of reasoning.

The \emph{Web Ontology Language} (short: \emph{OWL}) is a specification by the W3C with the purpose of representing knowledge in machine-parsable ways.
In the OWL~1 Web Ontology Language Guide~\citep{smith_owl_2004}, the W3C explains that OWL~1 can be divided into its three sublanguages \emph{OWL~1~Full}, \emph{OWL~1~Lite} and \emph{OWL~1~DL}.
These sublanguages themselves can be further divided and differ in their expressiveness with OWL~1 Full being the feature-richest.
It can be seen as an extension of RDF, which includes meta-modelling capabilities of RDF Schema (short: \emph{RDFS})~\citep{mcguinnes_owl_2004}.
OWL~1~Lite and OWL~1~DL on the other hand can be understood as extensions of a subset of RDF~\citep{mcguinnes_owl_2004}.
Ontologies created using OWL~1~Lite are subsets of those built with OWL~1~DL which themselves are subsets of OWL~1~Full ontologies.
Detailed differences can be taken from the OWL~1 semantics specification~\citep{patel-schneider_owl_2004}.

With rising amount of features, the complexity of reasoning increases as well:
Reasoning in OWL~1~Lite is complete for \textsc{ExpTime}, in OWL~1~DL for \textsc{NexpTime} and OWL~1 Full is even undecidable~\citep{horrocks_reducing_2004}.
The revision of OWL~1, called OWL~2, even extends the feature set of the sublanguages, resulting e.g.\ in a reasoning complexity of \textsc{2NexpTime} for OWL~2~DL~\citep{grau_owl_2008}. 
Thus, especially when reasoning on billions of triples, it is important to decide on using the least complex sublanguage that offers all needed features for the given modelling purpose.

Commonly used for modelling knowledge while keeping reasoning feasible are sublanguages of OWL~2~DL, also referred to as \emph{profiles}.
These profiles are characterised by further restricting the feature set and hence making reasoning more feasible~\citep{ter_horst_completeness_2005,krotzsch_owl_2012}. 
Hereby, these ontology languages also become simpler to implement, extend and easier to understand~\citep{krotzsch_owl_2012}. 
The W3C distinguishes three OWL~2 profiles, namely \emph{OWL~2~QL}, \emph{OWL~2~RL} and \emph{OWL~2~EL}~\citep{motik_owl_2012}.
These profiles have a reduced reasoning complexity of as low as \textsc{PTime} for OWL~2~RL~\citep[cf.][]{motik_owl_2012,cao_web_2011}. 
One of the restrictions that all the profiles share is that for axioms which define subclass inclusion, they disallow unions of classes.
Otherwise, reasoning on these ontology languages would become \textsc{NP}-hard already.
Similarly, all profiles forbid the use of negations and universal quantifiers on the left-hand side.
Their concrete restrictions are explained in the OWL~2 profile specification~\citep{motik_owl_2012} and are outside of this paper's scope.
However, \citet{krotzsch_owl_2012} summarises their most important characteristics as follows:

OWL~2~QL was designed as a query language to ease information retrieval from ontologies as a database.
Hence, queries may extract matching data together with facts inferred from it in \textsc{LSpace}~\citep{motik_owl_2012}.
This sublanguage's design enables \emph{query rewriting}, where the query is rewritten e.g.\ into an SQL query that can be directly executed on an SQL database~\citep{bishop_implementing_2011}. 
The technique of query rewriting is discussed in more detail in \autoref{s:queryRewriting} with an approach being presented where the query is split up into related rules that eventually may return instances.

OWL~2~RL is often applied to contexts where the TBox is notably smaller than the ABox, which is common for the \semweb{}. 
It was developed as a scalable solution and allows querying on big datasets while retaining most of OWL~2~DL's expressiveness~\citep{motik_owl_2012}.

OWL~2~EL is likewise aimed at scenarios with big ABoxes~\citep{motik_owl_2012}.
Thus, it is used e.g.\ for biomedical ontologies like Systematized Nomenclature of Medicine Clinical Terms\footnote{\url{http://www.ihtsdo.org/} --- last accessed 06 March 2015, 16:30} (short: \emph{SNOMED CT}), which gathers information about human diseases.
It is the most restricted profile and therefore the least expressive.

Besides the OWL~2 profiles, \emph{OWL~pD*} (also known as \emph{OWL~Horst}) is also often used for reasoning on huge amounts of data.
It is an extension of RDFS combined with a subset of OWL~1~\citep{ter_horst_completeness_2005}.
As a result, it becomes more expressive than RDFS while remaining less computationally complex than OWL~1~Full~\citep{urbani_owl_2010}. 
Reasoning in OWL~pD* has a complexity of \textsc{NP} and \textsc{P} for some special cases~\citep{ter_horst_completeness_2005}.
\citet{urbani_owl_2010} describe it as \enquote{a de facto standard for scalable OWL reasoning}~\cite[][p.216]{urbani_owl_2010}. 
The development of OWL~2~RL was partly influenced by OWL~pD*~\citep{motik_owl_2012}. 
Thus, both are frequently used for large-scale reasoning~\citep[e.g.][]{bishop_implementing_2011,hogan_saor_2010}. 

Regardless of the revision of OWL, it can be said that OWL~Lite is not expressive enough for modelling most knowledge on the \semweb{}.
OWL~DL and OWL~Full however are not tractable for reasoning on huge datasets.
Therefore, restricted sublanguages are selected for knowledge modelling and reasoning on the \semweb{}.
OWL~pD* and OWL~2~RL are the languages most commonly applied to this task, often using a materialisation strategy presented later in~\autoref{s:materialisation}.
Nevertheless, other profiles such as OWL~2~QL are also used in some reasoners following a query rewriting strategy as further explained in~\autoref{s:queryRewritingActual}.

\section{Approaches for Large-Scale OWL Reasoning}\label{s:techniques}
First of all, this section gives an overview of description logics, focussing on one concrete language as an example.
The description logics syntax and semantic will be used when subsequently presenting the two inference methods \emph{forward chaining} and \emph{backward chaining}.
The later sections go into detail about implementations of these methods.
\autoref{s:related} will furthermore give an overview of a selection of OWL and/or \semweb{} reasoners.

Description logics are a first-order logic subset and allow the modelling of knowledge~\citep{baader_description_2003}. 
They are the foundation of OWL and other ontologies.
A common example of description logics is the \emph{Attributive Language with Complements} (short: $\mathcal{ALC}$)~\citep{schmidt-schausz_attributive_1991}
It consists of \emph{concepts}, \emph{individuals} and \emph{roles}.
A concept can e.g.\ be a \textsf{Human} or \textsf{Building}.
Individuals can be seen as elements of concepts e.g.\ \textsf{Peter} or \textsf{EiffelTower}.
A function---also called an \emph{interpretation}---associates the individuals with concepts, so that for example \textsf{Peter {:} Human} expresses that the individual \textsf{Peter} satisfies the \textsf{Human} concept.
Examples for roles are \textsf{hasChild} or \textsf{builtBy}.

To link these elements with each other, $\mathcal{ALC}$ introduces \emph{connectives}.
One such connective is the concept inclusion ``$\sqsubseteq$'', which allows expressing that one concept is more general than the other.
For example \textsf{Woman $\sqsubseteq$ Human} may model that the gender neutral concept of a human includes women.
Like in set theory, concepts can also be unified so that e.g.\ \textsf{Woman $\sqcup$ Man $\sqsubseteq$ Human}.
The intersection of concepts indicated by ``$\sqcap$'' is included analogously.
There is also a top concept ``$\top$'' which is the most universal concept and thus subsumes all others.
Likewise, the bottom concept ``$\bot$'' models the concept of nothingness.
An example, using the negation connective ``$\neg$'', would be \textsf{Woman $\sqcap\ \neg$Woman $\sqsubseteq \bot$}.
Existential and universal quantifiers---``$\exists$'' and ``$\forall$'' respectively--- allow expressing knowledge including roles.
For example, ``a building is built only by humans or no one'' can be modelled as \textsf{Building $\sqsubseteq\ \forall$builtBy.Human}.
Similarly, modelling a parent as someone who has had a child can be expressed with \textsf{Parent $\sqsubseteq \exists$hasChild.$\top$}.
The given introduction to $\mathcal{ALC}$ and description logics in general is incomplete and the interested reader may consult \citet{schmidt-schausz_attributive_1991} and \citet{baader_description_2003} respectively.

Forward and backward chaining differ in the \enquote{direction} of reasoning.
Reasoning is used to derive implicit knowledge from ontologies by applying terminological knowledge to the explicitly modelled data~\citep{baader_description_2003}.
Forward chaining is \emph{data-driven} meaning that reasoning will start from existing data and infer new knowledge as long as it is possible~\citep{shi_scalable_2014}. 
Given are two example class subsumption rules in \autoref{eq:example_rule1} and~\ref{eq:example_rule2}.
\begin{equation}\label{eq:example_rule1}
  X \sqsubseteq P
\end{equation}
\begin{equation}\label{eq:example_rule2}
  Z \sqcup P \sqsubseteq Y
\end{equation}
A forward chaining approach searches for rules with matching antecedents and then assumes true consequents.
\autoref{eq:forward} shows how example data may express $a$ being of class $Z$ and $X$.
Next, it can be matched against the antecedents of the example rules and hence reasoned to being of class $Y$.
\begin{equation}\label{eq:forward}
  a:Z \sqcup a:X
  \xRightarrow{(1)} a:Z \sqcup a:P
  \xRightarrow{(2)} a:Y
\end{equation}

Backward chaining on the other hand is \emph{goal-driven}~\citep{shi_scalable_2014}. 
It divides a goal into smaller subgoals and tries to resolve those.
By matching rules for true consequents and then assuming true antecedents, backward chaining reasons for data matching the initial goal.
Likewise, \autoref{eq:backward} illustrates a backward chaining application assuming $a$ being of class $Y$.
\begin{equation}\label{eq:backward}
  a:Y
  \xRightarrow{(2)} a:Z \sqcup a:P
  \xRightarrow{(1)} a:Z \sqcup a:X
\end{equation}

The following sections explain how the approaches can be used for \semweb{} reasoning and what advantages they have over the other.
\autoref{s:materialisation} presents a common way to use forward chaining for this task by also presenting a typical programming model and an overview of an implemented reasoner.
Subsequently, \autoref{s:queryRewriting} discusses the application of backward chaining to reasoning on billions of triples.
Analogously, the main properties and approaches of an implementation are shown.

\subsection{Materialisation}\label{s:materialisation}
Materialisation is a forward chaining approach~\citep{kiryakov_owlimpragmatic_2005}. 
The idea behind is to compute all inferences prior to reasoning and storing them for later querying~\citep{shi_scalable_2014}. 
In this section, after discussing the most important advantages and disadvantages to this technique, a common programming model called \emph{MapReduce} is explained.
Furthermore, a reasoner using MapReduce is presented.

Computing all inferences first once, allows fast query answering as it is comparable to lookups in a database.
On the other hand, the initial materialisation process is time and memory consuming.
As an example, \texttt{owl:sameAs} is one of the most commonly found axioms according to \citet{hogan_rdfs_2013}.
It is used to express equivalent individuals.
If a reasoner would be na\"{\i}vely implemented, a full closure would be in $\mathcal{O}(n^2)$.
On a corpus containing $33,052$ equivalent individuals, \citet{hogan_scalable_2012} prospected $1,092,434,704$ triples and an additional two billion for those individuals being included in other statements. 
Another downside is that materialisation must be done anew every time the data is updated.
Due to the nature of the \semweb{}, data may change frequently and hence regular updates are necessary to ensure recent results.

\subsubsection{MapReduce}
MapReduce is a programming model developed by \citet{dean_mapreduce_2004} for Google Inc.\ with the goal to process big amounts of data efficiently.
It tries to achieve this by allowing distributed and parallel data processing and thus reducing the load on single machines/cores.
The most frequently used implementation is Apache Hadoop\footnote{\url{https://hadoop.apache.org/} --- last accessed March 08 2015, 17:30}.
MapReduce can also be used for reasoning on \semweb{} data \citep[cf.][]{tachmazidis_large-scale_2012,urbani_scalable_2009} for which it then becomes a forward chaining approach.
The process is split into three consecutive steps as described by \citet{dean_mapreduce_2004}:
\begin{enumerate}
  \item \textbf{Map} is the first phase and describes the act of pre-processing input data as a list of key-value pairs.
    The result of every processed key-value pair is immediately emitted.
    Thus, this process may be parallelised and can hence speed up data processing.
    Its method signature can be given as:
    \begin{itemize}
      \item[] $\mathtt{Map}\left(k_1,v_1\right) \rightarrow \mathtt{list}\left(k_2,v_2\right)$
    \end{itemize}
    The different indices express potentially different datatypes.
  \item \textbf{Shuffle} collects all results of the first phase and groups them by their key.
    Hence, the datatypes remain the same and a method signature can be given as:
    \begin{itemize}
      \item[] $\mathtt{Shuffle}\left(\mathtt{list}\left(k_2,v_2\right)\right) \rightarrow \mathtt{list}\left(k_2,\mathtt{list}\left(v_2\right)\right)$
    \end{itemize}
  \item \textbf{Reduce} is the phase in which the values with the same key are being processed.
    Values are being merged together and returned as a smaller result set.
    To avoid having to load the complete data into memory, the reduce phase is only given an iterator over the values.
    The list of returned values is of the same datatype as the intermediate values returned by the mapping phase.
    Its signature can be given as:
    \begin{itemize}
      \item[] $\mathtt{Reduce}\left(k_2,\ \mathtt{list}\left(v_2\right)\right) \rightarrow \mathtt{list}\left(v_2\right)$
    \end{itemize}
    However, the commonly used implementation Apache Hadoop is more liberal concerning the domain of return values, by having the following method signature~\citep{apache_software_foundation_apache_2014}:
    \begin{itemize}
      \item[] $\mathtt{Reduce}\left(k_2,\ \mathtt{list}\left(v_2\right)\right) \rightarrow \mathtt{list}\left(k_3,v_3\right)$
    \end{itemize}
    Similar to the map phase, the reduce phase is generally executable in parallel.
\end{enumerate}

\begin{figure}
  \includegraphics[width=\linewidth]{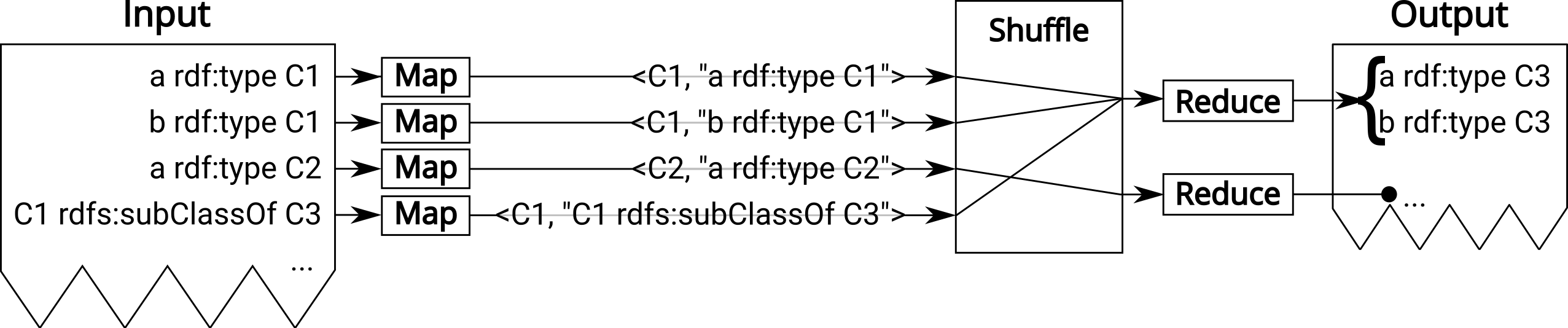}
  \caption{An example MapReduce process for the rule $s\ \mathtt{rdf:type}\ x,\ x\ \mathtt{rdfs:subClassOf}\ y\ \Rightarrow\ s\ \mathtt{rdf:type}\ y$. The figure was taken from \citep{urbani_scalable_2009} and modified to explicitly include the shuffle phase.}
\label{f:mapreduce}
\end{figure}

\autoref{f:mapreduce} illustrates a MapReduce application on an ontological rule.
In this example, the two triples of the rule's antecedent share the common variable $x$.
Hence, it is selected as key for the map phase.
After this pre-processing, the shuffling groups all results by key.
The reduce phase then produces the output drawn from the intermediate results matched against the consequent.
At this point, the full reasoning process may not already be finished.
The reason for this is that the inferred knowledge matches part of the rule's antecedent anew.

MapReduce applied to reasoning is not a trivial task which needs only one run as illustrated by the rule in \autoref{f:mapreduce}.
Instead, there are multiple difficulties when trying to implement this approach in a way that is relatively light on space and time consumption.
A few of these are presented in the next section together with an example MapReduce algorithm.

\subsubsection{WebPIE as an Example Implementation of MapReduce}
\citet{urbani_webpie_2012} employ MapReduce in their reasoner \emph{Web-scale Inference Engine} (short: \emph{WebPIE}) for reasoning on OWL~pD* rules.
The authors discuss the shown problem as a \emph{fixed-point iteration}, expressing that the MapReduce process must be repeated until no more new triples are returned.
This results in an additional problem as duplicates may be generated by having to apply some rules multiple times.
Besides trying to generate as few duplicates as possible throughout the reduce phase, an additional MapReduce step is added that solely matches duplicates and removes them~\citep{urbani_scalableMA_2009}. 
In the map phase, the step iterates over all triples and returns these triples as key, which removes duplicates in the succeeding shuffle phase.
Then, in the reduce phase, triples are only emitted if they were inferred to distinguish them from the original input triples.

In general, \citet{urbani_scalable_2009} try to execute as few MapReduce steps as possible.
Yet, the processing of some rules together with others is either not possible or would have a heavy impact on performance.
Thus, the authors implement various MapReduce steps which focus on certain rules.

One, which will be discussed as an example, is the algorithm shown in \autoref{alg_mapReduce}.
It calculates subclass relations and is separated into a map and a reduce method as shown in \autoref{l_map} and \autoref{l_reduce} respectively.
The algorithm operates on triples having either \texttt{rdf:type} or \texttt{rdfs:subClassOf} as predicate.
Depending on their predicate, the map phase uses different flags as keys---in this case either \texttt{0} or \texttt{1}---together with the triple's subject.
The value associated with the key is the input triple's object.
Considering an example input triple \texttt{\_x rdfs:subClassOf \_y}, then the output's key flag will be \texttt{1}, its subject \texttt{\_x} and the value \texttt{\_y}.

The reduce method first removes all duplicates within the values.
For the sake of simplicity, these values will be referred to as classes.
In Lines~\ref{l_mem1} and~\ref{l_mem2} all TBox triples matching the classes as a subject are loaded into memory.
These correspond to all the classes' superclasses.
Continuing the earlier example, having a triple \texttt{\_y rdfs:subClassOf \_z} in the TBox results in \texttt{\_z} being added to the list of superclasses.

The reduce method distinguishes the input values based on their formerly associated predicate in the original triple as indicated by the key's flag.
In both cases, the method iterates over all superclasses.
It checks that the currently processed superclass is not already contained in the set of classes to prevent duplicates.
Whenever this is ensured, the method returns a new triple associated with a key.
As the key is irrelevant in this scenario, the pseudocode reduce method returns \texttt{null}.
The triple models that the original input's subject is either of the same type as a matching superclass or also a subclass of that superclass.
Thus, inferred triples will be returned that respect the class hierarchy.
In the formerly given example, this would be \texttt{\_x rdfs:subClassOf \_z}.

\begin{lstlisting}[caption={The algorithm to do RDFS subclass reasoning initially presented by \citet{urbani_scalable_2009}. This listing features a corrected and simplified version including changes from \citet{urbani_scalableMA_2009}.},label={alg_mapReduce}]
map(key, value):#\label{l_map}#
  // key: source of the triple (irrelevant)
  // value: triple
  if (value.predicate = "rdf:type"):
    key.flag = 0
  if (value.predicate = "rdfs:subClassOf"):
    key.flag = 1
  key.subject = value.subject
  emit(key, value.object)#\DNumber#

reduce(key, iterator values):#\label{l_reduce}#
  // key: flag + triple.subject
  // iterator: list of classes
  values = values.unique // filter duplicate values#\label{l_duplicates}\DNumber#

  for (class in values):#\label{l_mem1}#
    superclasses.add(subclass_schema.getSuperclasses(class))#\label{l_mem2}\DNumber#

  switch (key.flag):
    case 0: // we're doing rdf:type
      for (superclass in superclasses):
        if !values.contains(superclass):
          emit(null, new Triple(key.subject, "rdf:type", superclass))
    case 1: // we're doing rdfs:subClassOf
      for (superclass in superclasses):
        if !values.contains(superclass):
          emit(null, new Triple(key.subject, "rdfs:subClassOf", superclass))
\end{lstlisting}

\citet{urbani_webpie_2012} present approaches for tackling the difficulties they faced throughout building WebPIE\@.
However, their solutions are mostly tightly coupled with the ruleset of OWL~pD*.
Nevertheless, they also propose two more generic optimisation strategies.

As the TBoxes are commonly a lot smaller than the ABoxes when reasoning with data from the \semweb{}, the TBoxes can often times be fully loaded into memory.
This was also done in the algorithm shown in~\autoref{alg_mapReduce}.
The resulting advantage is that then the triples of instances can be streamed and thus directly processed.
But, \citet{urbani_webpie_2012} note that this is not possible for rules which require joins between multiple instance triples in their antecedent.
\autoref{e:horstSameAs} shows such a rule from the OWL~pD* ruleset.
\begin{equation}\label{e:horstSameAs}
  v\ \mathtt{owl:sameAs}\ w,\ w\ \mathtt{owl:sameAs}\ u \Rightarrow v\ \mathtt{owl:sameAs}\ u
\end{equation}
There, for both triples in the antecedent, matching instances in the ABox must be looked up each.
Again, the authors propose a solution concretely fit to the affected OWL~pD* rules.

However, they also describe a common technique for reducing the overhead created by rules considering the \texttt{owl:sameAs} axiom as in \autoref{e:horstSameAs}.
The input is modified so that synonymous instances are replaced by a unique identifier representing an entire equivalence class each \citep[e.g.][]{urbani_owl_2010,bishop_owlim_2011}.
As a result, the required space and computation time are both reduced drastically~\citep{liu_large_2011}.

Although having only highlighted a few bottlenecks, it should have become clear that applying MapReduce to reasoning on billions of triples is a complex task.
Albeit there being some reoccurring problems shared by most OWL sublanguages, many optimisation approaches are tailored to concrete rules.
Therefore, there is a lack of efficient universal solutions.

\subsection{Backward Chaining}\label{s:queryRewriting}
With backward chaining, reasoning is done at runtime, once the query is posed.
Hence, no prior computation is needed.

Backward chaining has two advantages over materialisation as described by \citet{urbani_querypie_2011}.
First, there is no need for precomputation due to the runtime reasoning.
Neither is there generally a need for computing a full closure because any reasoning solely needs to be done as far as it is required to answer the query.
As one result, an application exclusively using backward chaining may instantly be usable without prior time- and space-consuming computation.

Second, the results to a query consider recent data modifications.
That is, when changes happen to the data such as deletions or additions, they are instantly retrievable through the reasoning done by query rewriting.
Furthermore, this means that after changes have been applied to the data, no computationally and memory-intense recomputation has to be done.

However, backward chaining also faces a great disadvantage compared with materialisation.
As reasoning must be done with every new query, answering these is often time-consuming.
Whereas for materialisation query matching instances may directly be returned from the fully reasoned knowledge base. 

\subsubsection{Query Rewriting}\label{s:queryRewritingActual}
When using backward chaining for reasoning on ontologies, some form of query rewriting is used.
The concept of query rewriting is to reformulate the query into a query that respects the ontology's terminological axioms and retrieves the matching instances~\citep{imprialou_benchmarking_2012}. 
There are various options concerning how a query can be transformed such as logical rewriting~\citep{gottlob_ontological_2011} or even concrete SQL query rewriting~\citep{bishop_implementing_2011}.
Both are actively used in the context of \emph{ontology-based data access} (short: \emph{OBDA}).

OBDA describes the idea of storing ABoxes in traditional databases like relational database management systems while allowing the use of ontologic constraints~\citep{gottlob_ontological_2011}.
Here, OWL~2~QL is often the preferred language as it was designed implicitly with OBDA as intended purpose~\citep{motik_owl_2012,kikot_owlql_2011}.
With OBDA, data access is offered through the ontology as an intermediate layer so that the queries are independent from the actual data storage~\citep{xiao_rules_2014}.
This allows a unified semantical access to different data sources.
Therefore, scalable, OBDA-enabled reasoning may become of special interest for \semweb{} reasoning due to its data being highly diverse.

Despite the existing potential, there are yet only few reasoners using query rewriting while scaling up to billions of triples.
There is especially a lack of reasoners that focus on implementing the concept of OBDA and pertain this scalability.
Thus, the subsequent section presents a reasoner that is not build around OBDA but instead uses an illustrative query rewriting approach while supporting reasoning on billions of triples.

\subsubsection{QueryPIE}\label{s:querypie}
\begin{figure}
  \includegraphics[width=\linewidth]{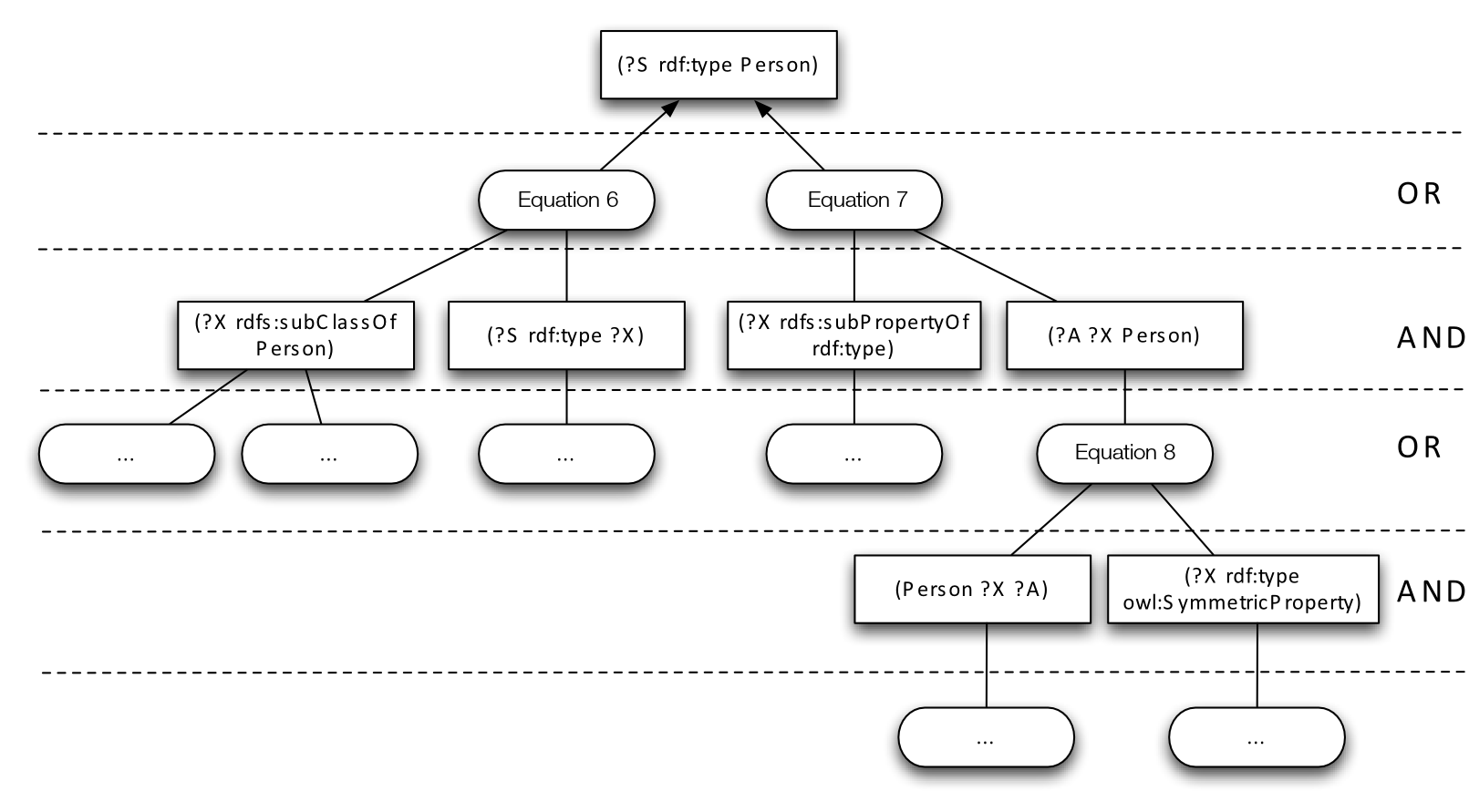}
  \caption{An example of a reasoning tree~\citep{urbani_querypie_2011} with the rule names being replaced to match this paper's equation numbers.} 
\label{f:queryRewriting}
\end{figure}

There is no prevalently used programming model for query rewriting like there is MapReduce for materialisation.
\citet{urbani_querypie_2011} do query rewriting for OWL~pD* by building a reasoning tree with the query being its root and the matching data its leaves.
In backward chaining fashion, the query---an input triple pattern---is matched against rule consequents and their antecedents will then be used as new query triples.
\autoref{f:queryRewriting} depicts a reasoning tree built from an example query \texttt{?S rdf:type Person}.
The branches containing rules are connected by logical \texttt{OR}s whereas the triples are connected with \texttt{AND}s.
This way, all instances from leaves may be returned that match the rule of their parent node.
However, it is not necessary for instance triples to match the rules of other branch nodes.
Thus all matching triples may be returned, disregarding through which rules they can be inferred.
In the given example, the consequents of Equations~\ref{e:r5} and~\ref{e:r4} match the input query.
\begin{equation}\label{e:r5}
  s\ \mathtt{rdf:type}\ x,\ x\ \mathtt{rdfs:subClassOf}\ y \Rightarrow s\ \mathtt{rdf:type}\ y
\end{equation}
\begin{equation}\label{e:r4}
  s\ p\ o,\ p\ \mathtt{rdfs:subPropertyOf}\ q \Rightarrow s\ q\ o
\end{equation}
Their antecedents are bound accordingly, substituting all variables where possible.
In the antecedent $s\ p\ o$ of \autoref{e:r4} for example, $o$ is bound to \texttt{Person}.
This allows to apply \autoref{e:o3} so that all variables from its antecedents are bound and reasoning can be continued on these.
\begin{equation}\label{e:o3}
  p\ \mathtt{rdf:type\ owl:SymmetricProperty},\ v\ p\ u \Rightarrow u\ p\ v
\end{equation}

The tree building finishes when each branch has been built.
A branch is built as deep as there are yet rules applicable or matching instances can be retrieved from the knowledge base.
This process may consume a lot of time and also space.
Thus, \citet{urbani_querypie_2011} propose two optimisation strategies:
\begin{itemize}
  \item[] \textbf{Precomputation} of frequently appearing branches.
  \item[] \textbf{Tree-pruning} describes the early discovery of branches which will not return any results and hence stopping to further follow them.
\end{itemize}

When implementing precomputation, its effectiveness is defined by how regular the selected branches appear.
\citet{urbani_querypie_2011} focus their optimisation on what they call \emph{terminological triple patterns}.
They define them as triple patterns whose object or predicate is a term from the RDFS or OWL vocabulary that they operate on.
An example is given in \autoref{e:terminologicalPattern}.
\begin{equation}\label{e:terminologicalPattern}
  \mathtt{?X\ rdfs:subPropertyOf\ rdf:type}
\end{equation}
Due to the commonly smaller TBox in \semweb{} data, there are few such terminological triple patterns.
Yet, they affect many queries.
As using forward chaining for precomputing would require the calculation of a full closure, backward chaining is applied on the selected terminological triple patterns.
For this, a tree is built bottom up by iterative backward chaining as illustrated in more detail in \citep{urbani_querypie_2011}.

The advantage of exclusively reasoning at runtime, is lost when using precomputation.
However, it is still a much faster process than full materialisation.
It especially remains fast enough to allow for frequent updates as typically found in \semweb{} data.
However, this approach can hence no longer be classified as pure query rewriting but is actually a hybrid approach~\citep{urbani_hybrid_2014}.

\vspace{1cm} 
\begin{lstlisting}[caption={The reasoner which operates on the precomputed terminological triple patterns as given by~\citet{urbani_querypie_2011} with small modifications to match this paper's pseudocode style.},label={alg_tiReasoner}]
reason(pattern):
  // Get rules where the query pattern is more specific than the rule's consequent:
  Rules applicableRules = ruleSet.applicable(pattern)#\DNumber#

  Results results = {}
  for (rule in applicableRules):
    antecedents = rule.instantiateAntecedents(pattern)#\label{l_bound}#
    // Perform reasoning to fetch all antecedents:
    for (antecedent in antecedents):
      if (antecedent != terminological)#\label{l_terminologicalIf}#
        // Recursive call to the reasoner:
        antecedents.add(reason(antecedent))
      antecedents.add(KnowledgeBase.read(antecedent))
    // Apply the rule using the antecedents triples:
    results += rule.applyRule(antecedents)#\DNumber#

  return results
\end{lstlisting}

\autoref{alg_tiReasoner} shows the algorithm in pseudocode used by \citet{urbani_querypie_2011} to reason after precomputing has been completed.
Hence, they call it the \emph{terminological independent reasoner}.
In the first step, all rules are retrieved whose consequent is more general than the query pattern.
This corresponds to the \texttt{OR}-levels of the reasoning tree.
While iterating over all rules, each rule's variables are subsequently bound to any matches from the query pattern as shown in \autoref{l_bound}.
In the second loop all antecedents of a rule are processed which corresponds to the \texttt{AND}-levels of the reasoning tree.
For this, the terminological independent reasoner is recursively called to further construct the branch.
Due to precomputing, this step is only needed if the currently processed antecedent is not terminological as checked in \autoref{l_terminologicalIf}.
Next, a lookup in the knowledge base is done which also includes the already computed terminological triple patterns.
After every branch of a rule have been computed, the rule is applied with all the previously computed antecedents.
In the end, all applied rules will be returned, meaning that all of their variables are bound to match the query pattern.

As a second optimisation strategy, tree-pruning is used.
It is built upon the formerly precomputed terminological triple patterns.
Whilst backward chaining, rules, whose antecedent matches such terminological triple patterns, are prioritised.
If these patterns do not return any instances, the rule will not be applied as they are more specific than the patterns. 
The processing of branches that are known not to return any results, will therefore be stopped early.
Thus, a lot of unnecessary reasoning can be prevented.

In the next section, a selection of reasoners using forward and/or backward chaining is presented.
However, all of the reasoners using backward chaining approaches either do not scale well or only partially support an OWL sublanguage.
QueryPIE is likely to be the first reasoner scaling to billions of triples while reasoning at runtime~\citep{urbani_hybrid_2014}.

\subsection{Other Reasoners}\label{s:related}
Besides WebPIE and QueryPIE, there are several other reasoners using forward and/or backward chaining.
This section gives an overview to a selection of available reasoners.

OWLIM~\citep{kiryakov_owlimpragmatic_2005} is a semantic repository allowing the storage of and reasoning on knowledge bases by employing full materialisation.
\citet{bishop_implementing_2011} state that OWLIM is capable to work on various rulesets including OWL~pD* and OWL~2~QL as well as OWL~2~RL\@.
OWLIM is divided into the free-for-use SwiftOWLIM and the commercial BigOWLIM~\citep{bishop_owlim_2011}. 
\citet{bishop_owlim_2011} state that SwiftOWLIM was developed for in-memory reasoning with smaller datasets.
According to the authors, BigOWLIM on the other hand was developed for reasoning on billions of triples.
For this, various optimisations were needed of which one is a special treatment of the \texttt{owl:sameAs} axiom.
BigOWLIM uses a canonical representation for each equivalence class~\citep{bishop_owlim_2011} similar to WebPIE's approach.
Furthermore, it employs a backward chaining approach on data deletion to prevent a new full materialisation~\citep{bishop_owlim_2011}.
According to \citet{bishop_owlim_2011}, its use allows BigOWLIM to be applicable to frequently updated data as is typical for data from the \semweb{}.

Another, yet non-commercial, reasoner supporting multiple profiles is TrOWL \citep{thomas_trowl_2010}.
It is an interface for multiple reasoners such as Quill or Pellet.
Besides the support of OWL~2~QL and OWL~2~EL there is also partial support for tractable reasoning with OWL~2~DL~\citep{pan_reasoning_2013}. 
Using reasoners like Pellet allows full OWL~2~DL reasoning support but reasoning is then no longer in polynomial time~\citep{dentler_comparison_2011}. 
Reasoning of OWL~2~QL is done by using backward chaining, namely query rewriting~\citep{thomas_trowl_2010}. 
OWL~2~EL on the other hand is reasoned in a forward chaining manner~\citep{pan_reasoning_2013}. 

\citet{tachmazidis_large-scale_2012} presented a reasoner which employs Apache Hadoop for materialisation of knowledge that uses rulesets implementing defeasible logic.
Their motivation for developing a reasoner on defeasible logic was to create inconsistency-tolerant reasoning that was able to deal with data of poor quality.
Although the reasoner does not operate on OWL but on RDF data, this concept may have a high relevance for the \semweb{} as data from different sources are likely to be of different quality.
Making use of parallelisation through MapReduce, \citet{tachmazidis_scalable_2012} were able to build a reasoner scaling to billions of triples.

\citet{stoilos_fuzzy_2005} developed a reasoner for an extension of OWL~pD* that supports fuzzy logic.
This allows expressing vagueness as already described in~\autoref{s:semweb}.
Their work also uses the MapReduce implementation of Apache Hadoop.
The authors followed the optimisation strategies of WebPIE and adapted them whenever necessary to support fuzzy logic.
As a result, \citet{stoilos_fuzzy_2005} claim to have created a fuzzy reasoner with a performance comparable to WebPIE\@.

Virtuoso Universal Server\footnote{\url{http://virtuoso.openlinksw.com/} --- last accessed 18 March, 22:00} is a Web server and triple store.
It is furthermore an OWL reasoner which allows backward as well as forward chaining~\citep{shi_scalable_2014,openlink_software_documentation_team_openlink_2014}. 
But, its support is restricted to a subset of OWL~2~RL~\citep{urbani_hybrid_2014}. 
Moreover, except for an RDF reasoner implementation~\citep{erling_towards_2008}, there do not seem to be any implementations targeting Web-scale triples.

Apache Jena\footnote{\url{https://jena.apache.org/index.html} --- last accessed 18 March, 08:30} is a framework written in Java to support building \semweb{} applications.
Its included OWL reasoner supports forward and backward chaining as well as a hybrid between the two~\citep{shi_scalable_2014}. 
However, throughout the research for this paper, no implementation using the Jena reasoner could be found which would scale to allow reasoning on billions of triples.

Another reasoner is F-OWL~\citep{zou_f-owl_2005} which primarily uses backward chaining.
The authors call their speed-up strategy \emph{tabling}.
Their approach is to store the results of already reasoned triples and look them up whenever possible.
As a result, the first queries are processed slowly but the system becomes increasingly faster on average for subsequent queries.
However, it only supports OWL~Full while neither being complete nor decidable.
According to its website\footnote{\url{http://fowl.sourceforge.net/} --- last accessed 17 March 2015, 22:00}, F-OWL has not been updated since 2003.
Additionally, according to its authors~\citep{zou_f-owl_2005} it does not scale and is thus unsuitable for reasoning on billions of triples.

\section{Conclusion}\label{s:conclusion}
The grand challenge of large-scale reasoning is to effectively use and reduce the time and space consumption.
Essentially, the approaches do not differ from regular reasoning methods.
Yet, the performance optimisation is crucial.

Additionally, the choice of ontology language may have heavy impact on the complexity of reasoning.
OWL~pD* and OWL~2~RL are commonly used for forward chaining approaches whereas OWL~2~QL is mainly used in backward chaining reasoners.
Furthermore, when reasoning on the \semweb{}, extensions of the languages can be of interest.
For example, using defeasible logic may allow reasoning with data of poor quality from different sources.
Likewise, OBDA may be a suitable technology for working and reasoning on \semweb{} data.

Reasoning on big knowledge bases modelled in OWL is still rarely done using backward chaining.
Instead, materialisation is the most common approach.
However, materialisation alone cannot fit the \semweb{}'s quickly changing nature.
Thus, future research should consider large-scale reasoning using a hybrid approach of backward and forward chaining.
Moreover, a requirement analysis of features needed for properly modelling and reasoning on \semweb{} data could be helpful.
Due to the diversity of data, a prior separation by topics would probably be beneficial.

\clearpage
\bibliographystyle{splncsnat}
\bibliography{./content/sources}
\end{document}